\pdfoutput=1
\documentclass{article}

\PassOptionsToPackage{square,numbers}{natbib}

\usepackage[final]{neurips_2022}

\usepackage[utf8]{inputenc} 
\usepackage[T1]{fontenc}    
\usepackage{hyperref}       
\usepackage{url}            
\usepackage{booktabs}       
\usepackage{amsfonts}       
\usepackage{nicefrac}       
\usepackage{microtype}      
\usepackage{xcolor}         

\usepackage[pdftex]{graphicx} 
\usepackage{subfig}
\usepackage{multirow}

\title{Parameter-Efficient Finetuning \\ of Transformers for Source Code}

\author{%
  Shamil Ayupov \\
  HSE University \\
  \texttt{shiayupov@edu.hse.ru} \\
  \And
  Nadezhda Chirkova$\bf{}^{}$\thanks{Now at Naver Labs Europe} \\
  HSE University \\
  \texttt{nchirkova@hse.ru} \\
  \\
}

\begin{document}

\maketitle

\begin{abstract}
  Pretrained Transformers achieve state-of-the-art performance in various code-processing tasks but may be too large to be deployed. As software development tools often incorporate modules for various purposes which may potentially use a single instance of the pretrained model, it appears relevant to utilize parameter-efficient fine-tuning for the pretrained models of code. In this work, we test two widely used approaches, adapters and LoRA, which were initially tested on NLP tasks, on four code-processing tasks. We find that though the efficient fine-tuning approaches may achieve comparable or higher performance than the standard, full, fine-tuning in code understanding tasks, they underperform full fine-tuning in code-generative tasks. These results underline the importance of testing efficient fine-tuning approaches on other domains than NLP and motivate future research in efficient fine-tuning for source code.
  
\end{abstract}

\section{Introduction}

The pipeline of pretraining Transformers on a large corpora and fine-tuning the pretrained model on a task-specific (downstream) dataset has become a dominating paradigm in natural language processing (NLP). Inspired by the success of pretrained Transfomers in NLP, recent studies adopt this approach to other data domains, e.g., to source code, and demonstrate its state-of-the-art performance in various code-processing tasks. 
For example, \citep{feng2020codebert, guo2020graphcodebert} show that the BERT model pretrained on source code corpora substantially outperforms earlier approaches  in code summarization (generating textual descriptions for code snippets), code search (retrieving code fragments given textual query) and code clone detection (deciding if a pair of code fragments is functionally identical). The works of \citep{wang2021codet5, ahmad2021unified, roziere2021dobf} further adapt pretraining approaches to source code by considering other Transformer-based models and proposing new pretraining objectives. 

Despite achiving better results in downstream tasks, pretrained models are known to suffer from the large model size. The widely used approach to mitigating this issue in NLP is parameter-efficient fine-tuning (PE fine-tuning), including such methods as Adapter Training, LoRA or prefix tuning \citep{houlsby2019parameter, hu2021lora, li2021prefix}. The general idea is to freeze the pretrained model and to introduce the small amount of additional parameters which will be fine-tuned for each task individually. Such approach substantially optimizes memory consumption at the deployment stage in case when one pretrained model is used in several applied tasks. It also reduces memory consumption at the fune-tuning stage because the gradients only need to be computed for additional task-specific parameters. However, the downstream performance of the parameter-efficient fine-tuning depends on the task: empirical studies show that PE fine-tuning approaches may outperform the standard (full) fine-tuning in the NLP tasks with small data but as the data size increases, full fine-tuning regains dominance \citep{he2021towards, chen2022revisiting}.

The large size of the pretrained Transformers especially matters in source code processing since it may prevent the pretrained models of code from deployment in integrated development systems (IDEs) and other developers tools. As an IDE includes modules for solving various code-processing tasks, it appears to be highly relevant to adopt PE fine-tuning approaches to code models. However, existing works mostly test PE fine-tuning on NLP tasks and, to the best of our knowledge, this approach was not previously tested on source code processing tasks, which are intuitively more complex. Moreover, source code data is often automatically collected from open-source repositories, e. g. by parsing functions and their descriptions from GitHub in code-to-text and text-to-code tasks. This results in a high level of noise in the training data (e.g. low-quality textual annotations, incorrect or buggy code) which makes it hard to apply the recommendations about applicability of the PE fine-tuning approaches in the small data setting.

In this work, we test two widely used PE fine-tuning approaches, Adapter Training and LoRA, and their combination for the pretrained models of source code. We find that PE fine-tuning cannot completely replace full fine-tuning. Particularly, in code-understanding downstream tasks,  the performance of PE fine-tuning is sometimes comparable or better than of full fine-tuning, but in more complex code-generative tasks, PE fine-tuning substantially underperforms full fine-tuning. These results underline the importance of testing PE fine-tuning approaches on other domains than NLP and motivate future research on efficient fine-tuning for source code. The source code of our experiments is available at \url{https://github.com/ShamerD/source-code-efficient-ft}.

\section{Methodology}

We consider two widely used PE fine-tuning approaches, Adapter Training (AT)~\citep{houlsby2019parameter} and LoRA~\citep{hu2021lora}, and also consider their combination. We outline the core ideas of the methods below and refer readers to the original papers for details. We vary the number of learnable parameters in these approaches and compare them with the full fine-tuning. 

\textbf{Adapter Training} (AT) relies on adding small fully-connected layers inside the Transformer. The layers are added after the Multi-Head Attention and Feed Forward layers. During fine-tuning, only these layers are trained while original model weights are frozen.
The drawback of this method is introducing inference latency as these layers themselves have computational costs.

\textbf{LoRA} introduces a learnable low-rank weights update $\Delta W = AB$ that will be summed up with the weights of the original model after fine-tuning: $W + \Delta W = W + AB$, where $W \in \mathbb{R}^{d \times k}, A \in \mathbb{R}^{d \times r}, B \in \mathbb{R}^{r \times k}, r < min(d, k)$. As a result, there is no additional inference latency.
The described modification can be applied to different parts of the model: to Query and Value weights in Multi-Head Attention layer (as in original paper) or to Feed Forward layer~\citep{he2021towards} (referred to as FF-LoRA in this work).

\paragraph{Models, tasks and fine-tuning data.}
We use CodeT5~\citep{wang2021codet5} and PLBART~\citep{ahmad2021unified} as the main models in our experiments since these are high-performing encoder-decoder models applicable to both generating and understanding tasks. We use the CodeT5 implementation for our experiments\footnote{\url{https://github.com/salesforce/CodeT5} (BSD-3-Clause license)}, 
employ HuggingFace's PyTorch Transformers library~\citep{wolf2020transformers} and use published pre-trained \verb|CodeT5-base| (223M) and \verb|PLBART-base| (140M) checkpoints in our experiments.

We consider four tasks from the CodeT5 and PLBART papers covering both code-understanding and code-generative scenarios. The considered datasets and metrics are widely used in source code processing and are a part of the CodeXGLEU benchmark~\citep{lu2021codexglue}.

\begin{figure*}[bthp]
    \centering
    \begin{tabular}{cccc}
        \subfloat[CodeT5: code summarization (Python) \label{fig:main:sum}]{\includegraphics[width=0.22\linewidth]{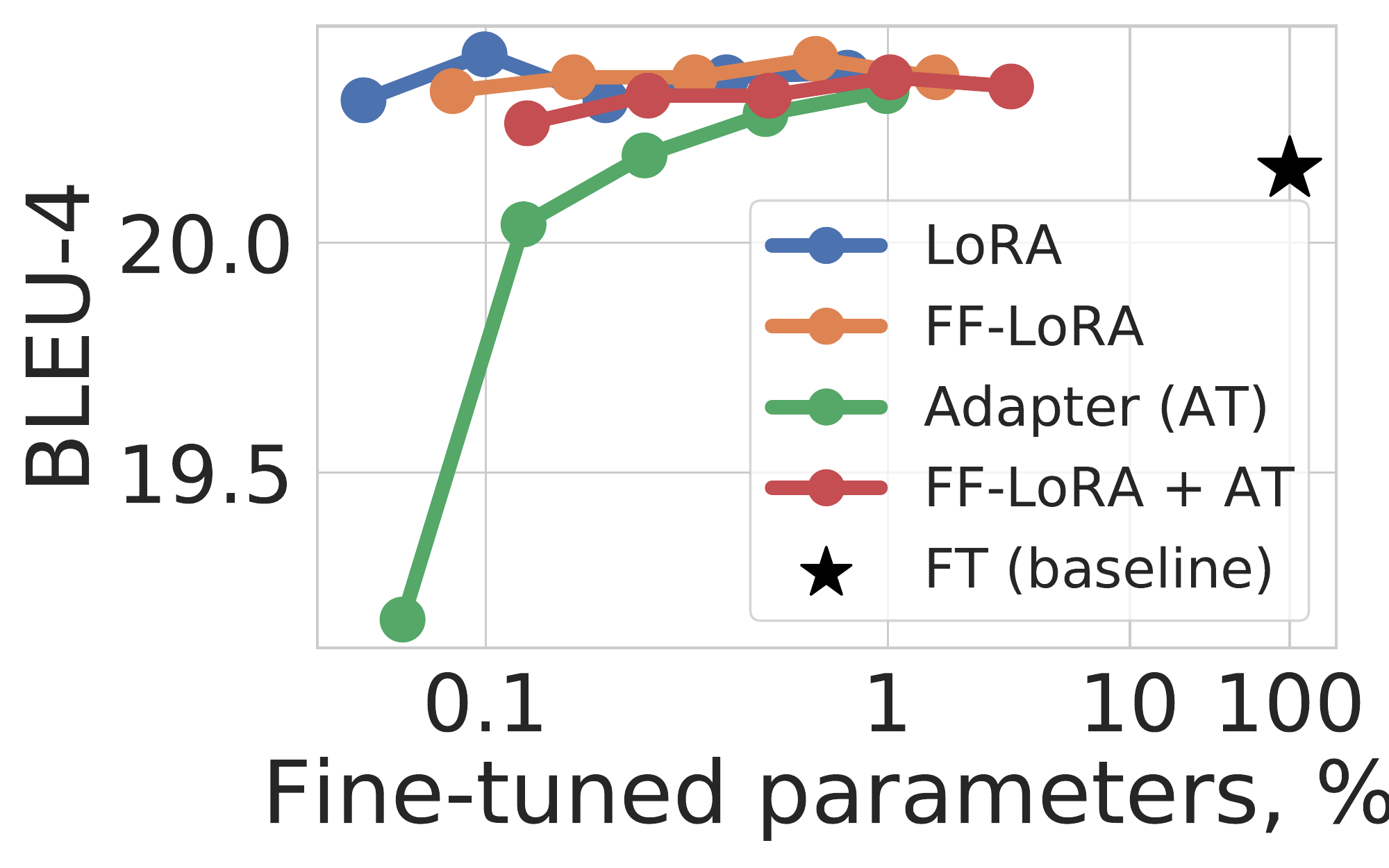}} &
        \subfloat[CodeT5: code clone detection (Java) \label{fig:main:clone}]{\includegraphics[width=0.22\linewidth]{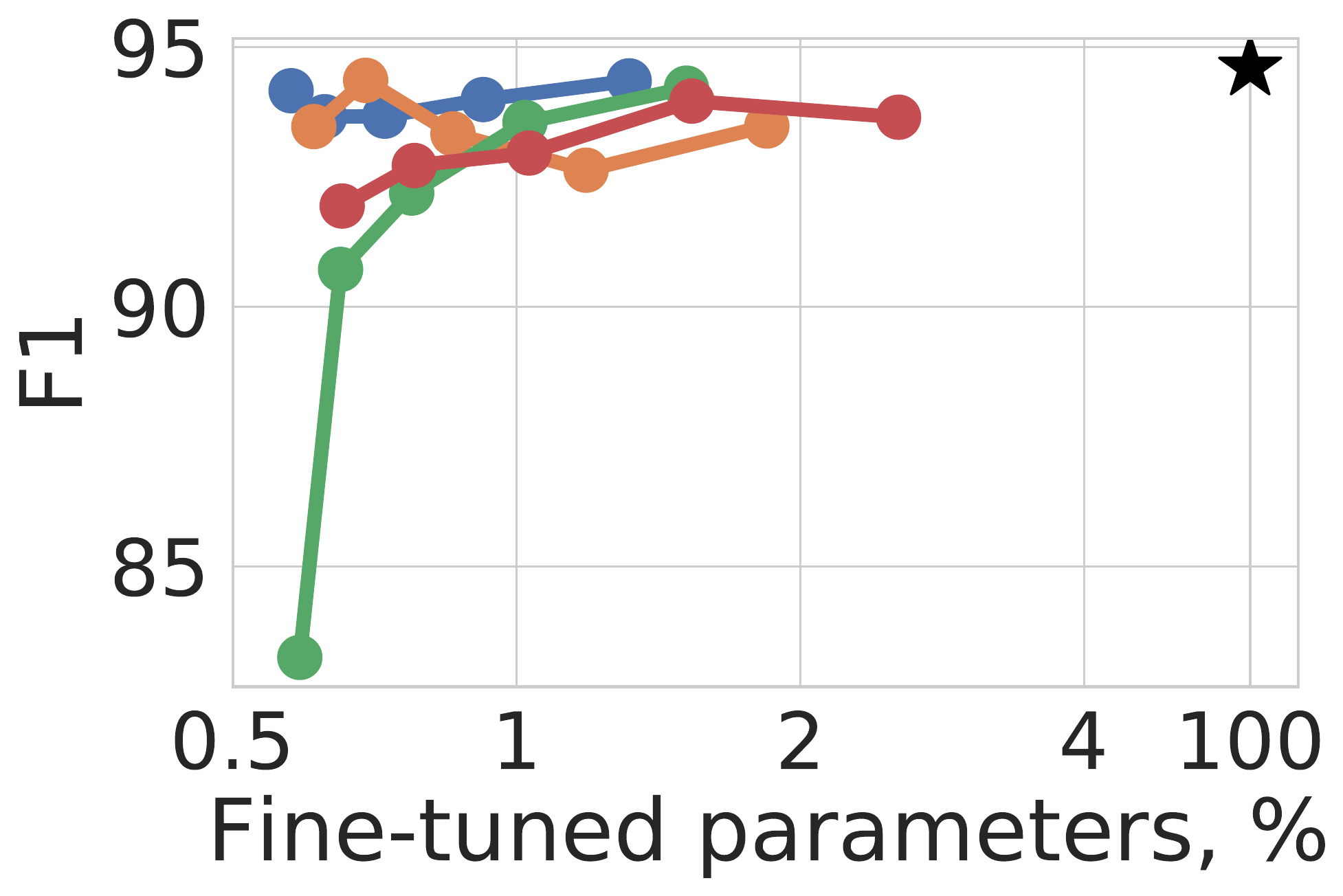}} &
        \subfloat[CodeT5: code generation (Java) \label{fig:main:gen}]{\includegraphics[width=0.22\linewidth]{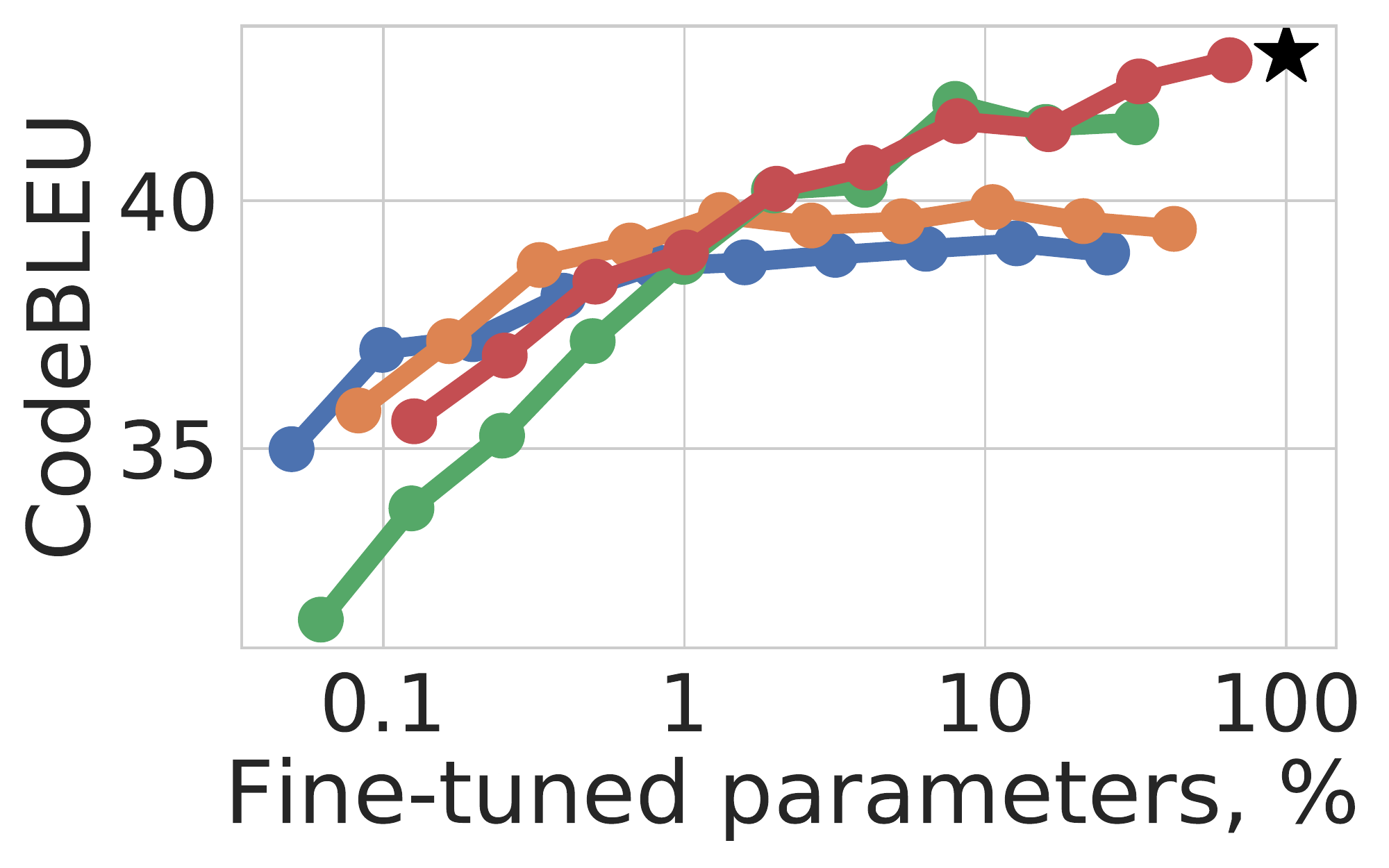}} &
        \subfloat[CodeT5: code translation (Java$\rightarrow$C\#) \label{fig:main:trans}]{\includegraphics[width=0.22\linewidth]{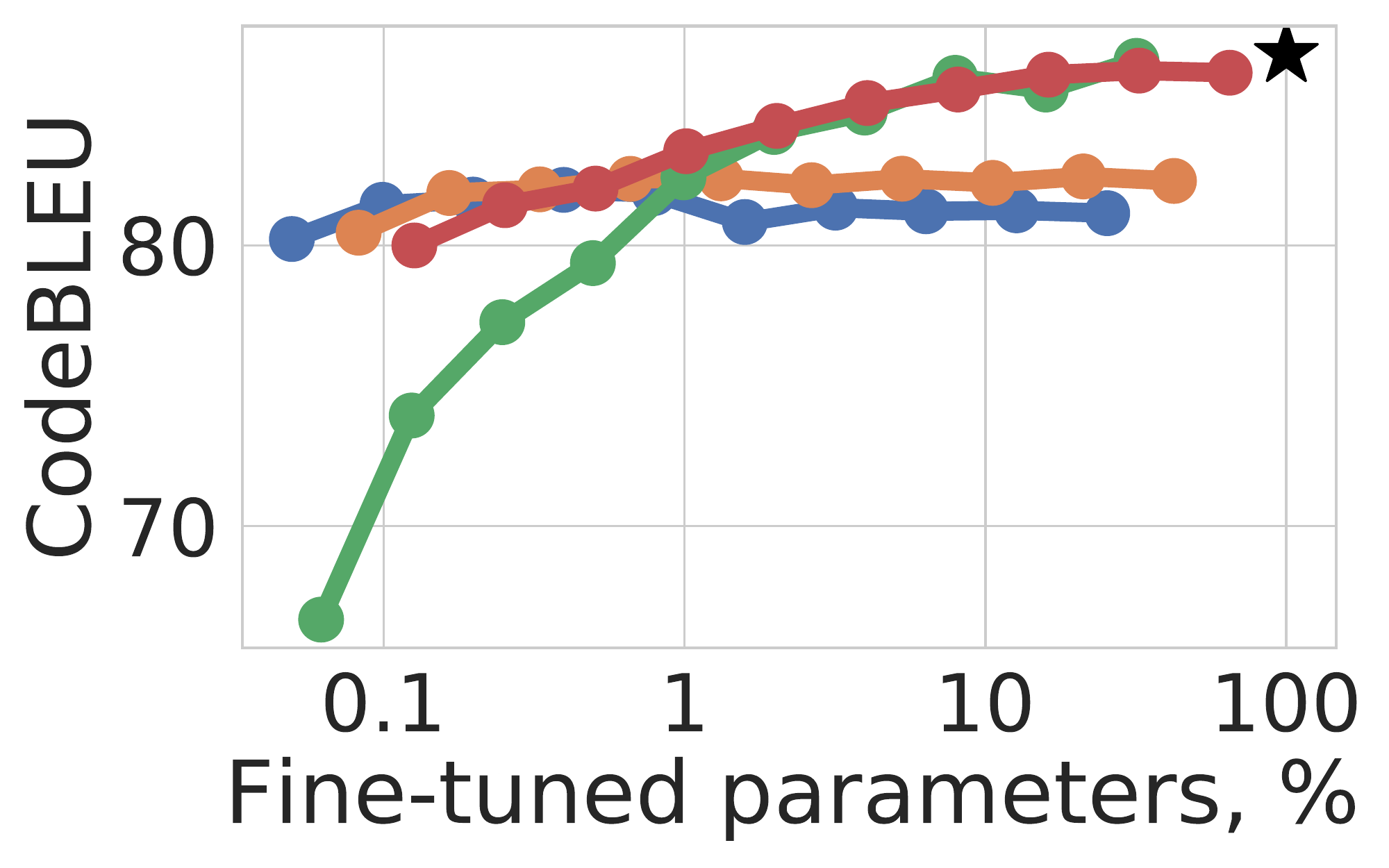}} \\
        \subfloat[PLBART: code summarization (Python) \label{fig:main:plsum}]{\includegraphics[width=0.22\linewidth]{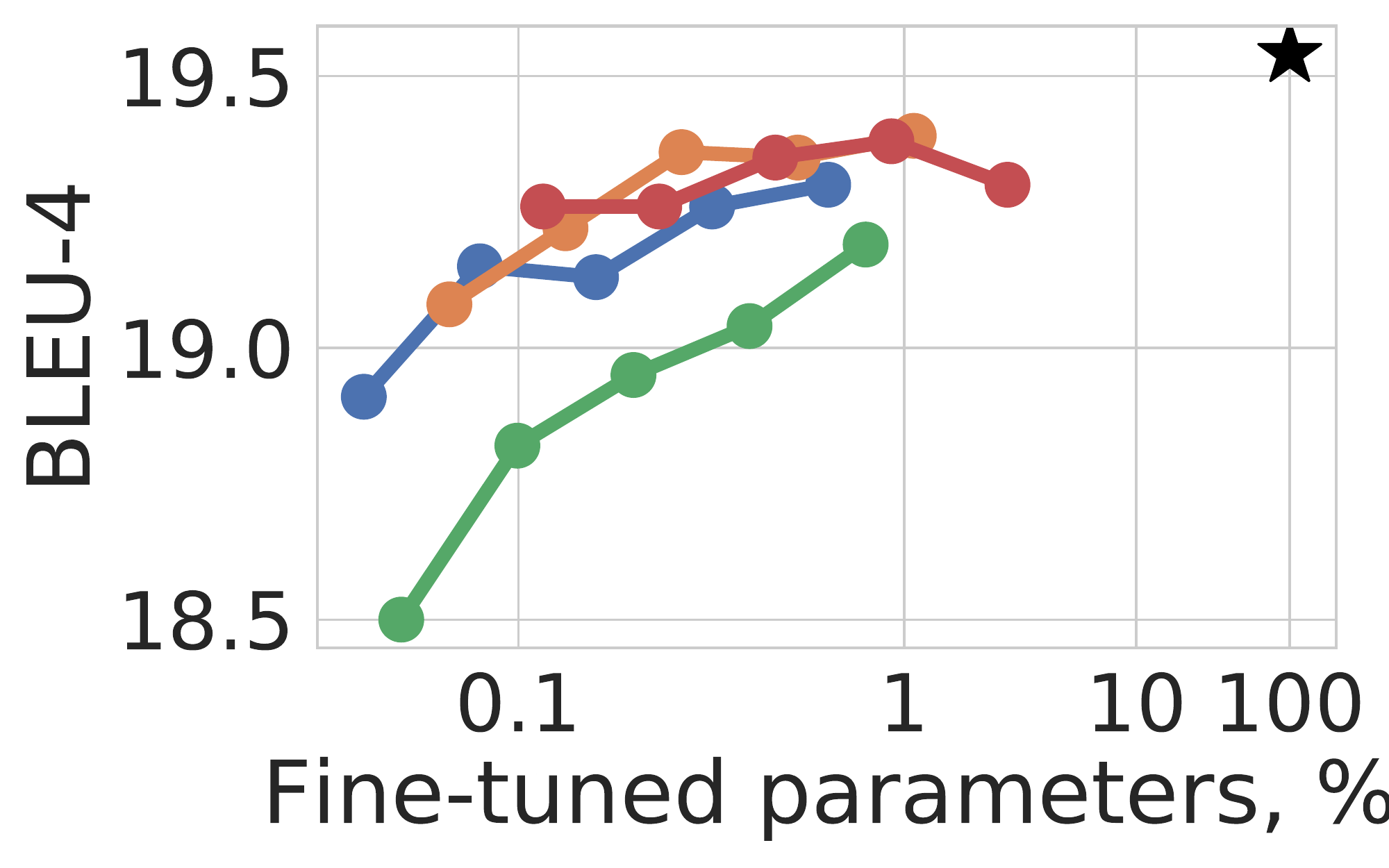}} &
        \subfloat[PLBART: code clone detection (Java) \label{fig:main:plclone}]{\includegraphics[width=0.22\linewidth]{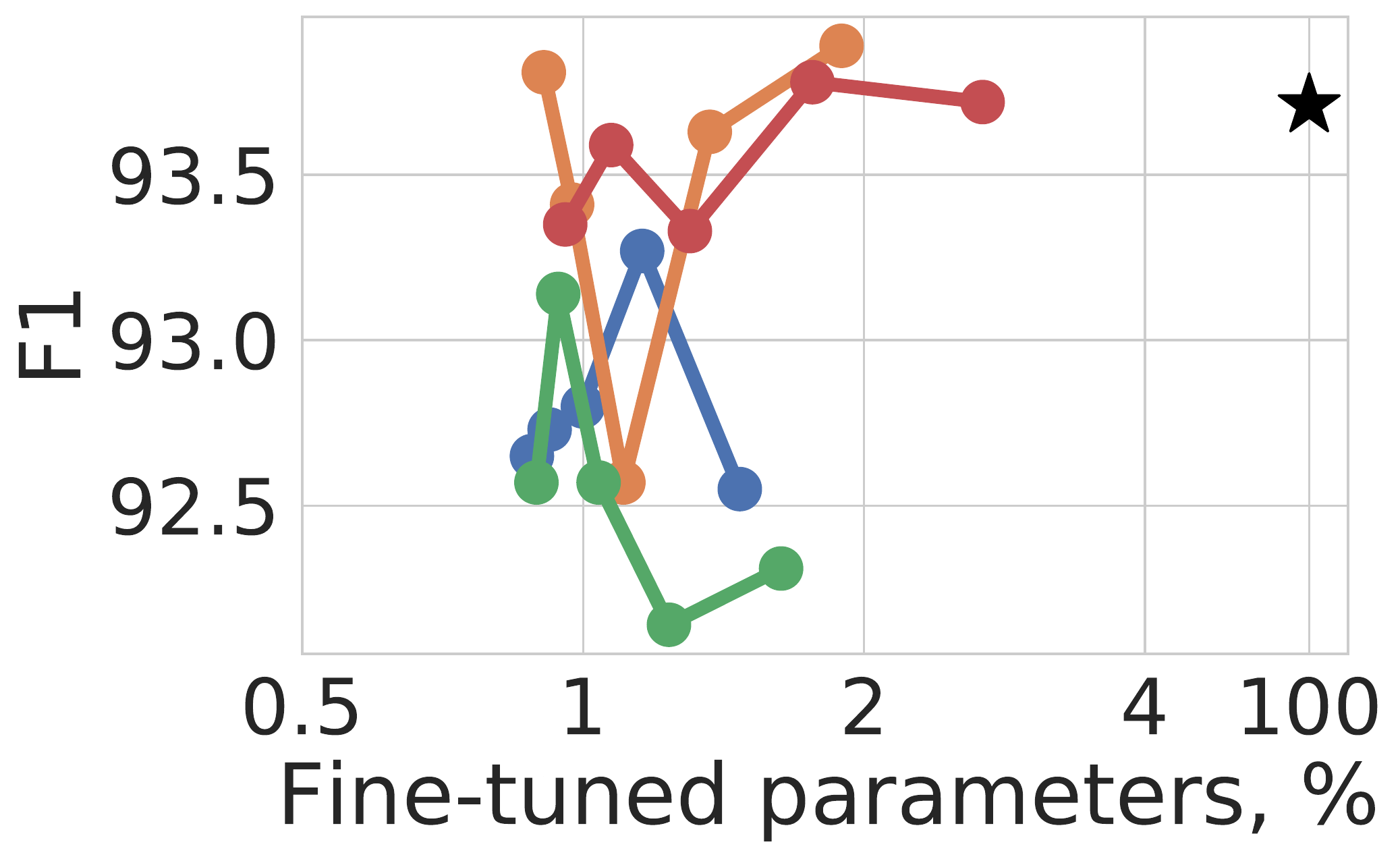}} &
        \subfloat[PLBART: code generation (Java) \label{fig:main:plgen}]{\includegraphics[width=0.22\linewidth]{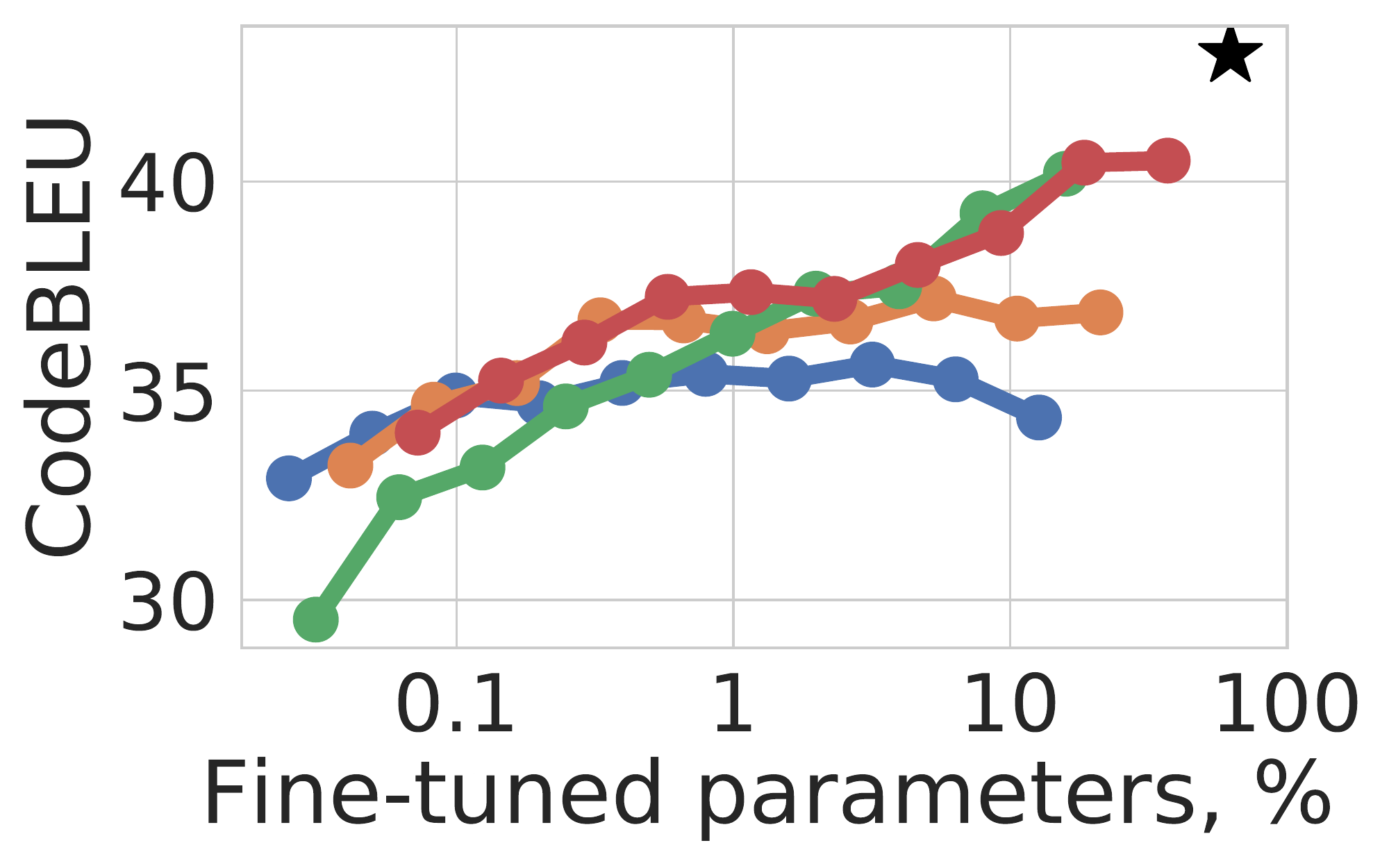}} &
        \subfloat[PLBART: code translation (Java$\rightarrow$C\#) \label{fig:main:pltrans}]{\includegraphics[width=0.22\linewidth]{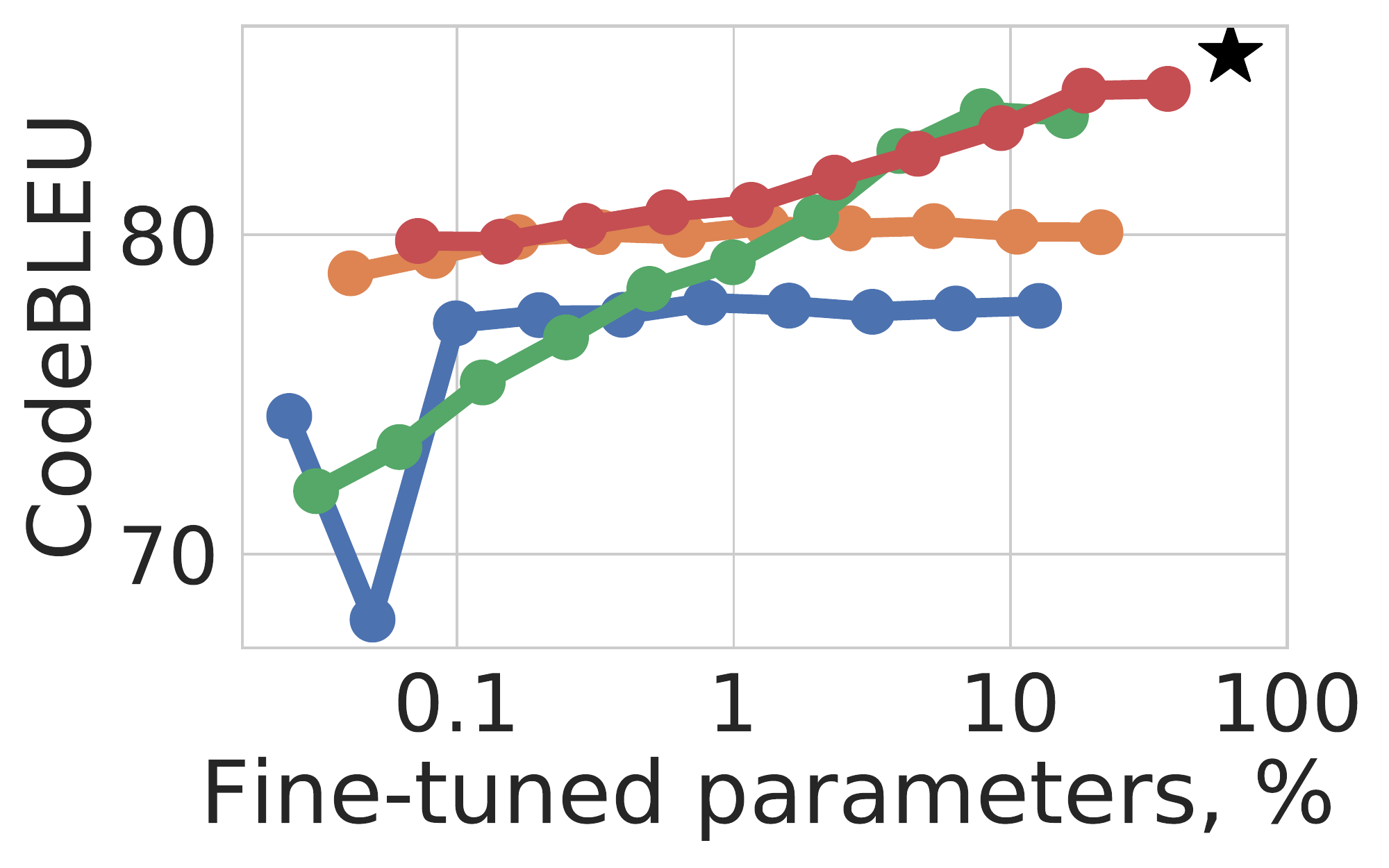}}
    \end{tabular}
    \caption{Quality of the PE fine-tuning methods in differents tasks on the test set vs. the number of trainable parameters. The numerical results are given in Appendix. }
    \label{fig:main_results}
\end{figure*}

\textbf{Code summarization} is the task to summarize code snippets into human-readable English descriptions. We consider Python and Go languages from the CodeSearchNet dataset~\citep{husain2019codesearchnet}. The dataset consists of 251,820~/~13,914~/~14,918 examples in Python and 167,288~/~7,325~/~8,122 examples in Go in the train~/~development~/~test sets respectively. This task is evaluated using smoothed \mbox{BLEU-4}~\citep{papineni2002bleu}.

\textbf{Code generation} is the task to generate code snippets given natural language (English) descriptions. The data consists of 100,000~/~2,000~/~2,000 (text + environment context, code) pairs on Java from the Concode dataset~\citep{iyer2018mapping}. This task is evaluated using CodeBLEU~\citep{ren2020codebleu}.

\textbf{Code translation} aims to translate code snippets from one programming language to another. The dataset consists of 10,300~/~500~/~1,000 code snippet pairs on C\# and Java languages and translation can be organized in both directions. This task is again evaluated using CodeBLEU. We additionally experiment with low resource setup (2.5k training examples) in Appendix \ref{sec:appendix}.

\textbf{Code clone detection} is the task of predicting whether two code snippets have the same functionality. The dataset is provided by \cite{wang2020detecting} and consists of 9,134 Java code snippets with the total of 901,028~/~415,416~/~415,416 labelled clone pairs. This task is evaluated using F1 score.

\paragraph{Experimental setup.} 
Fine-tuning hyperparameters were selected as in the original papers on CodeT5~\citep{wang2021codet5} and PLBART~\citep{ahmad2021unified}. 
The most important hyperparameter in the PE fine-tuning methods is the amount of learnable parameters which is determined by the inner layer size in AT and update rank in LoRA. We vary the latter hyperparameters in log-space (1, 2, 4, 8, 16). 
We report the quality on the test set, selecting the early stopping iteration in all experiments based on the quality on the development set.

\section{Experiments}

The main results for four tasks are shown in Fig. \ref{fig:main_results}, the results for additional programming languages are presented in Appendix. We find that in code-generative tasks (code generation and code translation, Fig. \ref{fig:main:gen}, \ref{fig:main:trans}, \ref{fig:main:plgen}, \ref{fig:main:pltrans}) the PE fine-tuning methods significantly underperform full fine-tuning and may achieve comparable performance only when the number of fine-tuned parameters approaches the number of all parameters in the model. 
In code understanding tasks (code summarization and code clone detection, Fig. \ref{fig:main:sum}, \ref{fig:main:clone}, \ref{fig:main:plsum}, \ref{fig:main:plclone}) the PE fine-tuning methods are sometimes able to achieve comparable or even better quality compared to full fine-tuning. Particularly, in clone detection, the PE fine-tuning of both models performs comparable to full finetuning, and in code summarization, this holds only for the CodeT5 model. However, the later result may be potentially attributed to the observation that the pretraining of publicly available CodeT5 checkpoint included the parallel part of the CodeSearchNet dataset, which is used for code summarization finetuning. To sum up, \textit{PE fine-tuning often substantially underperforms full fine-tuning in code-generative tasks and may achieve comparable or better results only in simpler code-understanding tasks.} 

\begin{figure}
    \centering
    \begin{tabular}{cc}
        \subfloat[Code summarization (Python, CodeT5) \label{fig:overfit:sum}]{\includegraphics[width=0.48\linewidth]{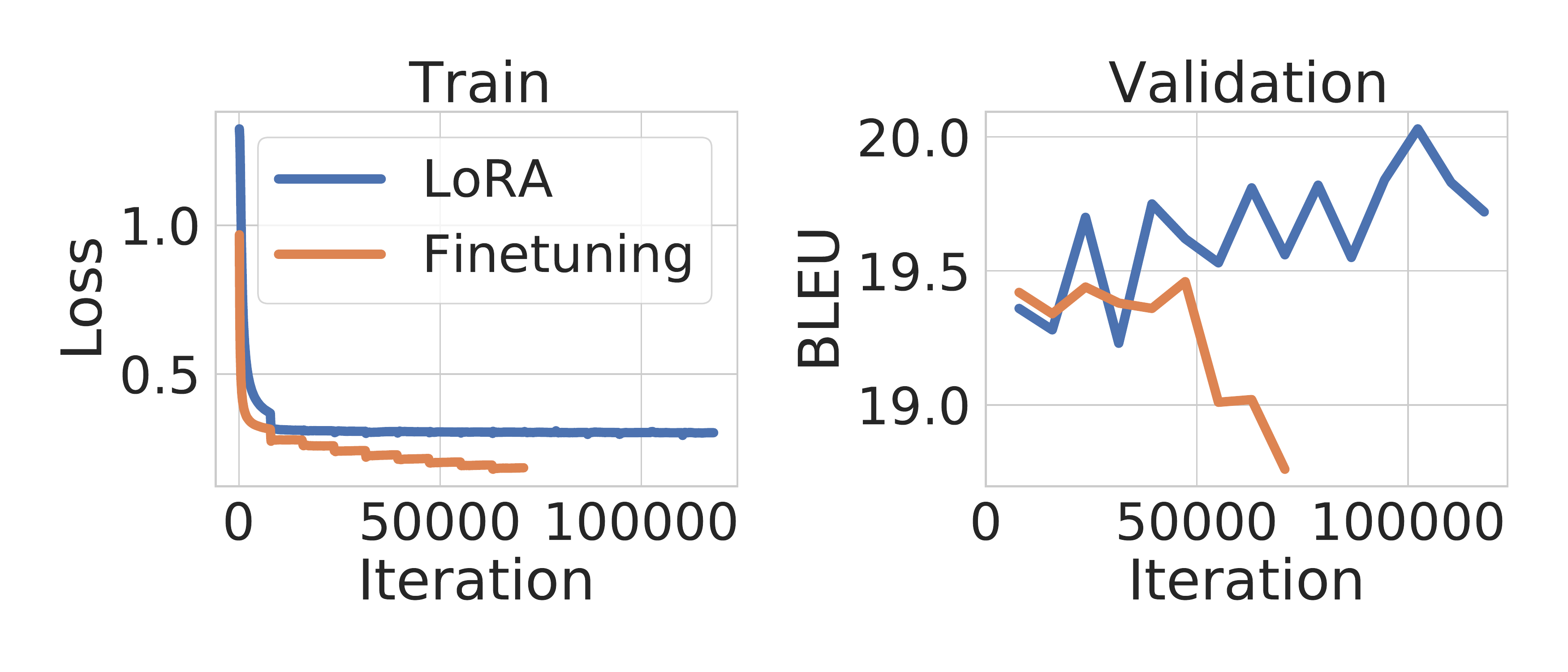}} &
        \subfloat[Code generation (Java, CodeT5) \label{fig:overfit:gen}]{\includegraphics[width=0.46\linewidth]{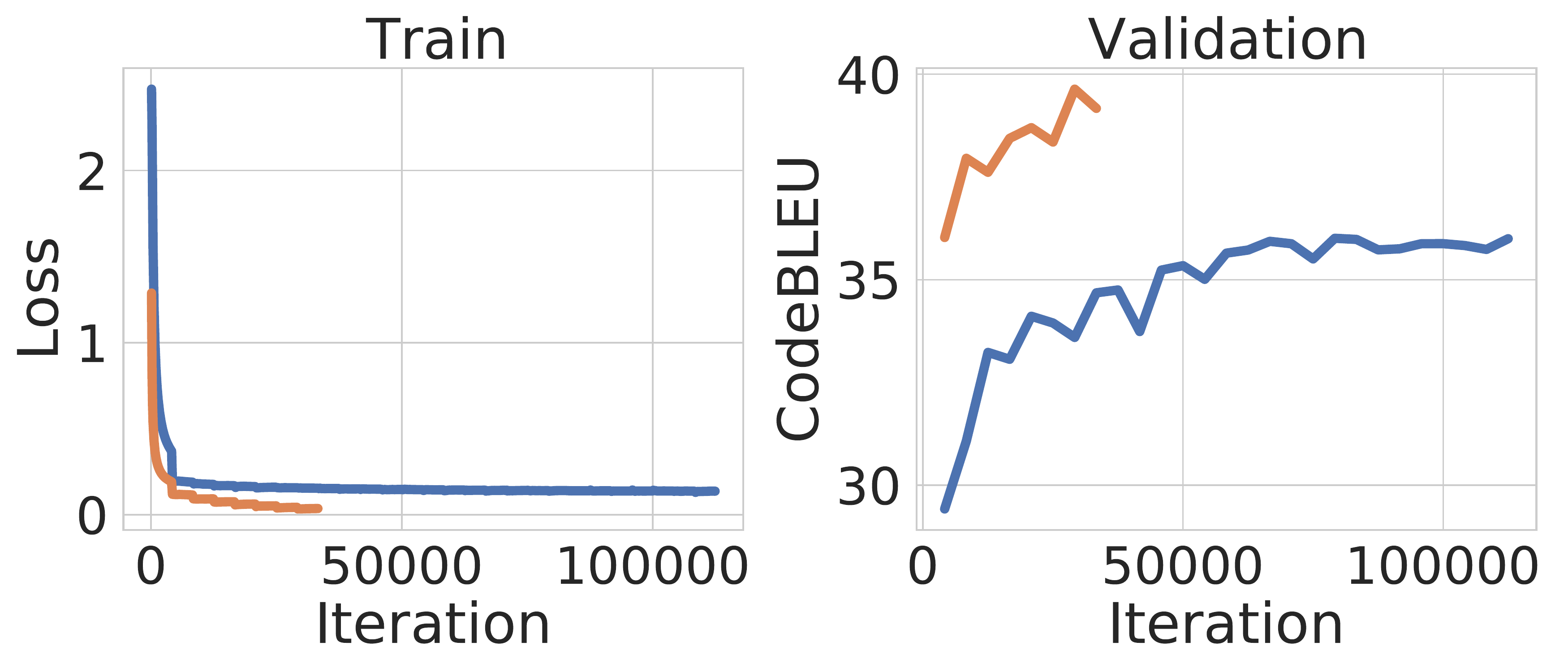}}
    \end{tabular}
    
    \caption{Example learning curves for two tasks.}
    \label{fig:learning_curves}
\end{figure}

We also analyse the training curves in Fig. \ref{fig:learning_curves} for both cases when PE fine-tuning underperforms full-finetuning and when they perform on par. This reveals that in the former case, the full fine-tuning severely overfits to the fine-tuning data (validation performance starts to decrease at some fine-tuning iteration) while PE fine-tuning acts like a regularizer. In the latter case, validation performance continues improving for both fine-tuning approaches.

Comparing different PE fine-tuning approaches, we find that Adapter Training is characterized by the monotonic dependence of the performance w.\,r.\,t. the number of fine-tuned parameters and underperforms LoRA~/~FF-LoRA for small numbers of fine-tuned parameters. However, in some tasks, Adapter Training outperforms LoRA~/~FF-LoRA for larger values of this hyperparameter. The combination of two approaches often behaves as an upper envelope of the corresponding quality plots, except the cases when Adapter Training performance is very low with the small number of fine-tuned parameters and reduces the combination's performance as well. FF-LoRA slightly outperforms LoRA or performs on par with it.

We also perform a qualitative analysis of the predictions on the code summarization task.
We inspected around 100 different models' predictions. Some can be seen in Table \ref{table:sumpy-2ex}, more can be seen in Appendix \ref{sec:appendix}.
The key observation is that models fine-tuned with the efficient fine-tuning methods tend to have shorter predictions based on a simple template.
At the same time the fully fine-tuned model outputs more elaborate predictions but they can contain irrelevant information or code fragments directly copied from the input.
This may also indicate that the full fine-tuning results in the model being overfit.

\section{Related Work}

Continuing the pioneer work of \citep{houlsby2019parameter, pfeiffer2020adapterfusion} on Adapter Training, recent research proposed a variety of PE fine-tuning approaches. Prompt-Tuning introduces prefix vectors (prompts) before the original input sequence \citep{li2021prefix} and updates only prompts during fine-tuning. Diff-Pruning \citep{guo2021parameter} learns sparse updates. LoRA \citep{hu2021lora} performs low-rank reparametrization of the update matrices. BitFit \citep{zaken2021bitfit} fine-tunes only bias vectors in Transformers. Authors of \citep{chen2022revisiting} conduct an empirical study of various methods for widely used NLP benchmarks and show the absence of the universal PE approach that would perform on par or better than full fine-tuning in all tasks, especially in the medium- and high-resource regimes. Our work extends the set of benchmarks on which PE fine-tuning was tested with source code processing tasks and comes to the similar conclusion.


\begin{table}
    \small
    \centering
    \begin{tabular}{||l|p{11cm}||}
        \hline
        Method & Example \\
        \hline
        GT & Scans through a string for substrings matched some patterns . \\*
        FT & matchall - Searches text for all patterns in patterns and returns a list of matches \\*
        LoRA & Return a list of all matching patterns in text . \\*
        FF-LoRA & Match all patterns in text . \\*
        AT & Match all patterns in text . \\
        \hline
        GT & int - > None Download a WHOLE course . Reuse the API call to save time . \\*
        FT & wanmen\_download\_by\_course\_topic and \\*
        {} & wanmen\_download\_by\_course\_part recursively \\*
        LoRA & Download wanmen by course . \\*
        FF-LoRA & Download wanmen files by course . \\*
        AT & Download wanmen by course . \\
        \hline
    \end{tabular}
    \caption{Examples of CodeT5 predictions when using different PE methods on code summarization task. GT stands for Ground Truth.}
    \label{table:sumpy-2ex}
\end{table}

\section{Conclusion}

In this work we studied the applicability of the efficient fine-tuning methods to Transformers for source codes in various source code processing tasks. Our main conclusion is that efficient fine-tuning often substantially underperforms full fine-tuning in source code processing tasks. This underlines that efficient fine-tuning approaches should be tested on other data domains except natural language and motivates the future development of efficient fine-tuning approaches for source code. As for practical recommendations, our experiments suggest using FF-LoRA or LoRA when the affordable number of learnable parameters is small and the combination of FF-LoRA and Adapter Training for larger parameter counts. 

\section*{Acknowledgments}
The work was supported by the Russian Science Foundation grant \textnumero 19-71-30020. 
The empirical results were supported through the computational resources of HPC facilities at HSE University~\cite{hpc}.

{
\small

\bibliography{custom}
\bibliographystyle{ieeetr}
}

\newpage
\appendix

\section{Appendix}
\label{sec:appendix}
\paragraph{Limitations and potential negative societal impacts}
To achieve a broader view on the considered problem, given limited resources, we selected a variety of applied tasks and programming languages, focused on two widely used PE fine-tuning approaches and considered two pretrained models. At the same time, given a wide range of PE fine-tuning approaches, pretrained models and applied tasks in the literature, it could be interesting to experiment with other models, tasks and approaches as well.


While the PE methods bring the main benefit to large models (with billions of parameters), they are not considered in this work due to limited computational resources. Experimenting with larger models could be an interesting direction for future resarch. 

Though we are not aware of potential negative societal impacts of our work, we note that the work may have potential negative environmental impact (about 300 Tesla V100 or A100 runs, each taking 8--24 hours, on the internal cluster).

\paragraph{Additional results}

\begin{figure*}[tbhp]
    \centering
    \begin{tabular}{ccc}
        \subfloat[CodeT5: code summarization (Go) \label{fig:add:sumgo}]{\includegraphics[width=0.3\textwidth]{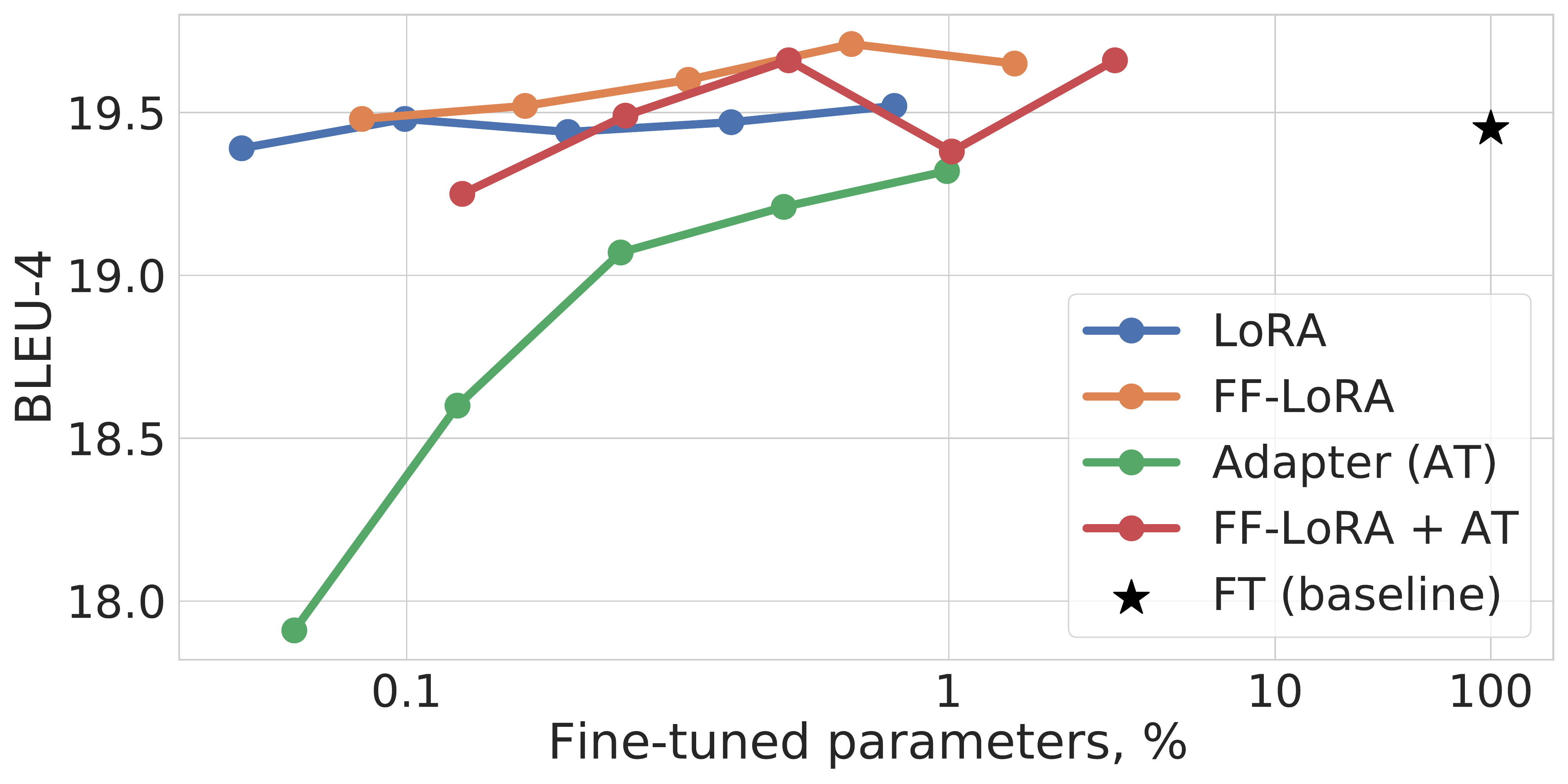}} &
        \subfloat[CodeT5: code summarization (Java) \label{fig:add:sumjava}]{\includegraphics[width=0.3\textwidth]{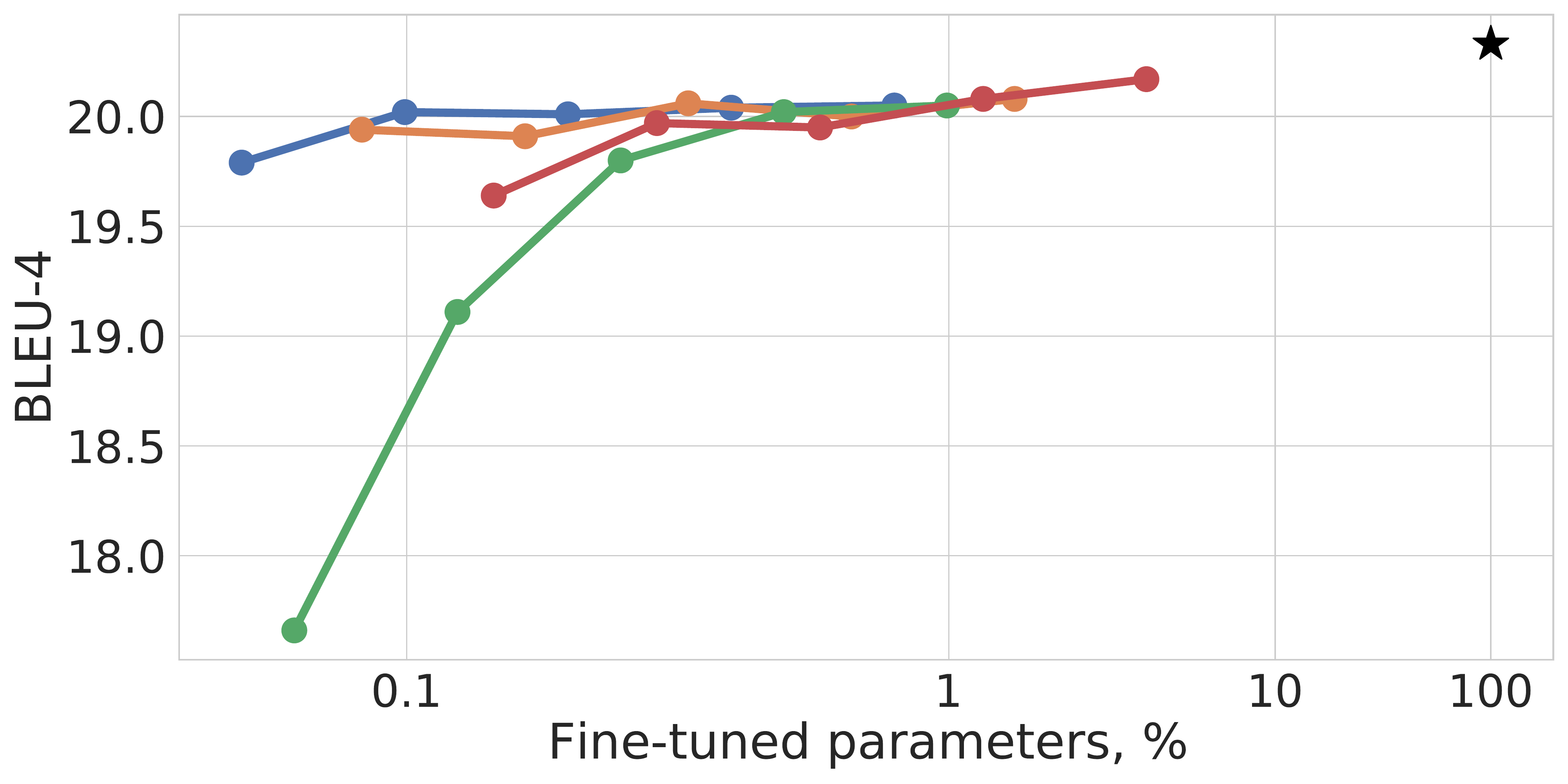}} &
        \subfloat[CodeT5: code translation (C\#$\rightarrow$Java) \label{fig:add:transcj}]{\includegraphics[width=0.3\textwidth]{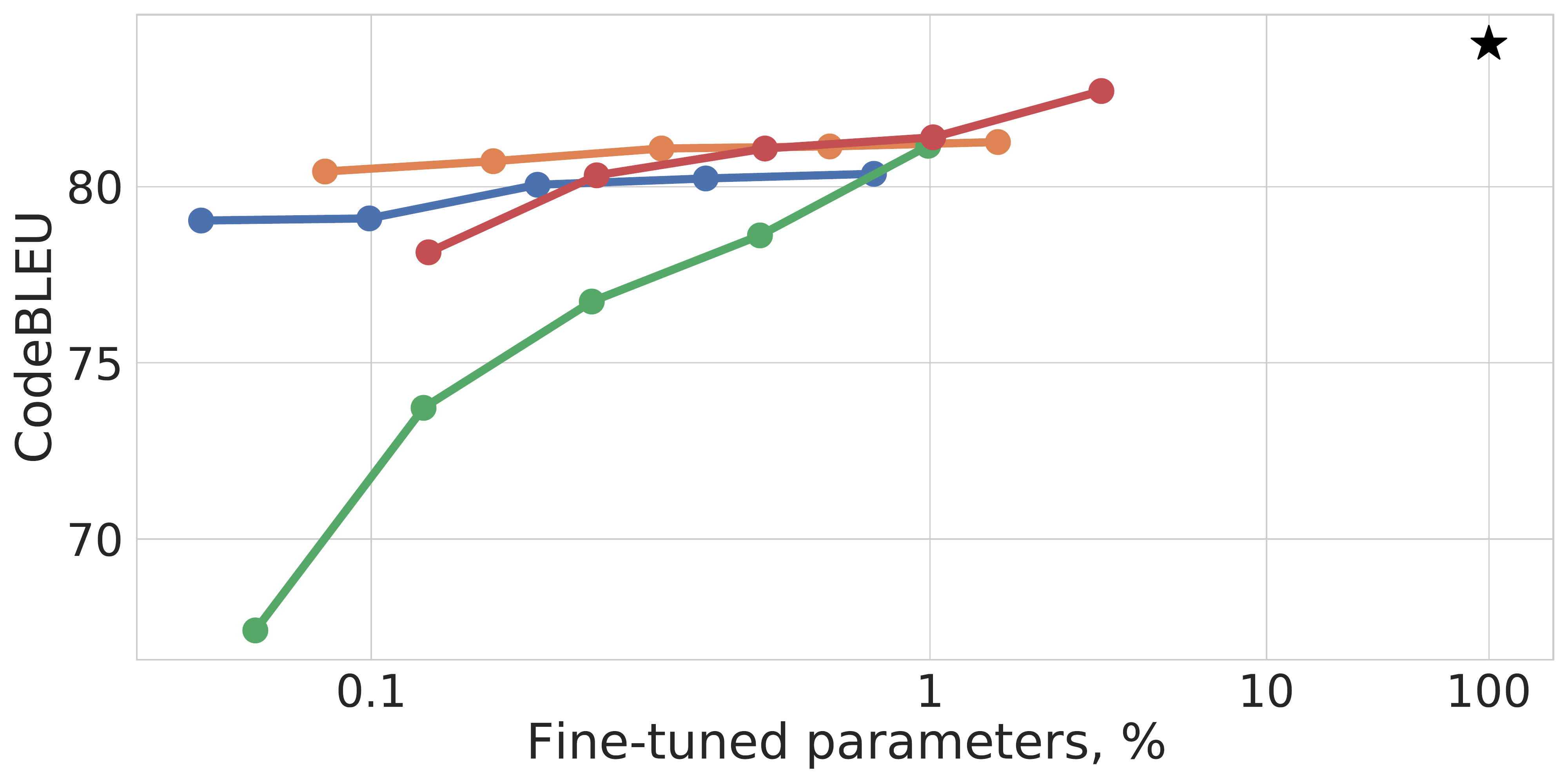}} \\
        \subfloat[PLBART: code summarization (Go) \label{fig:add:plbart_sumgo}]{\includegraphics[width=0.3\textwidth]{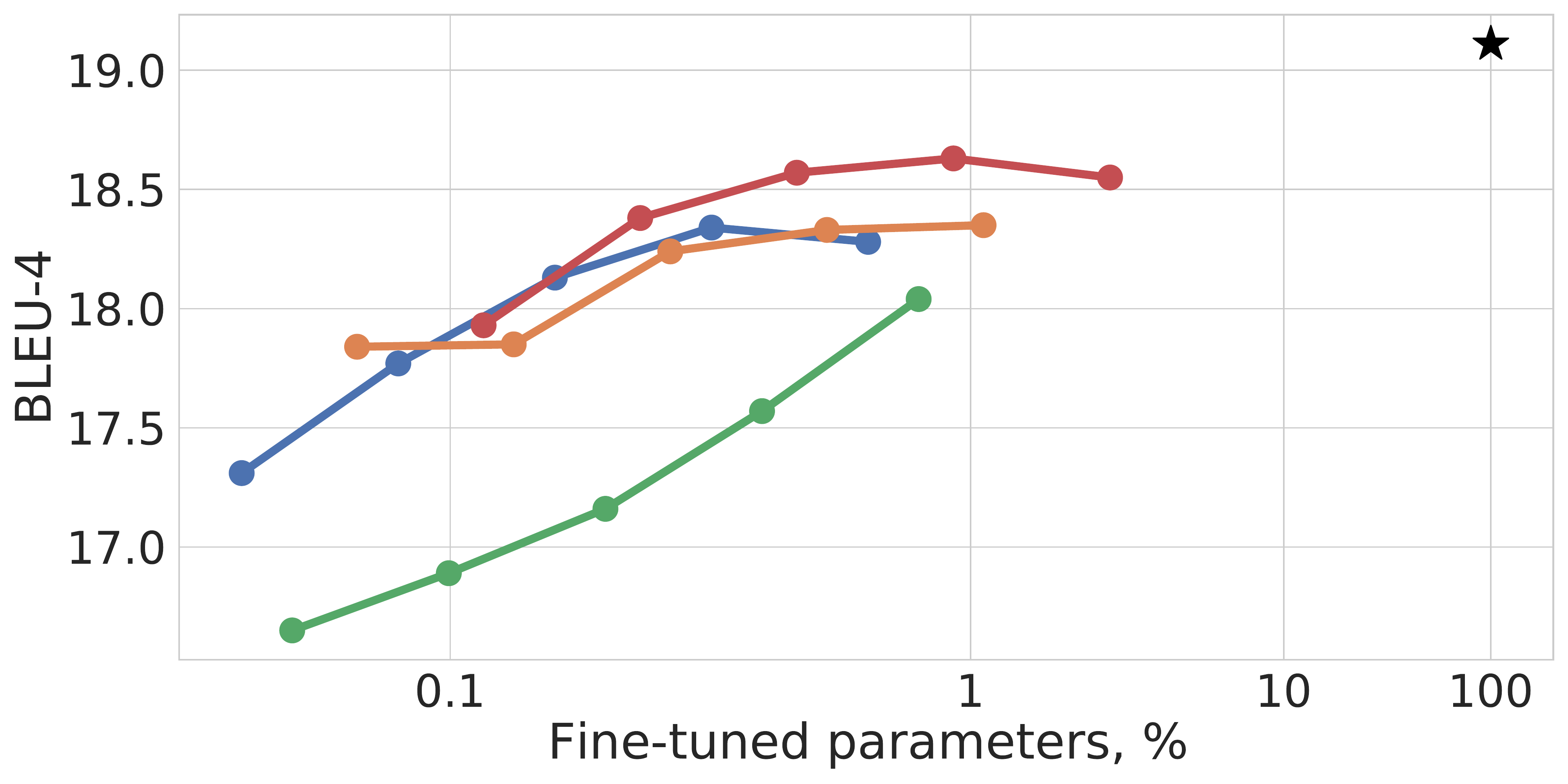}} &
        \subfloat[PLBART: code summarization (Java) \label{fig:add:plbart_sumjava}]{\includegraphics[width=0.3\textwidth]{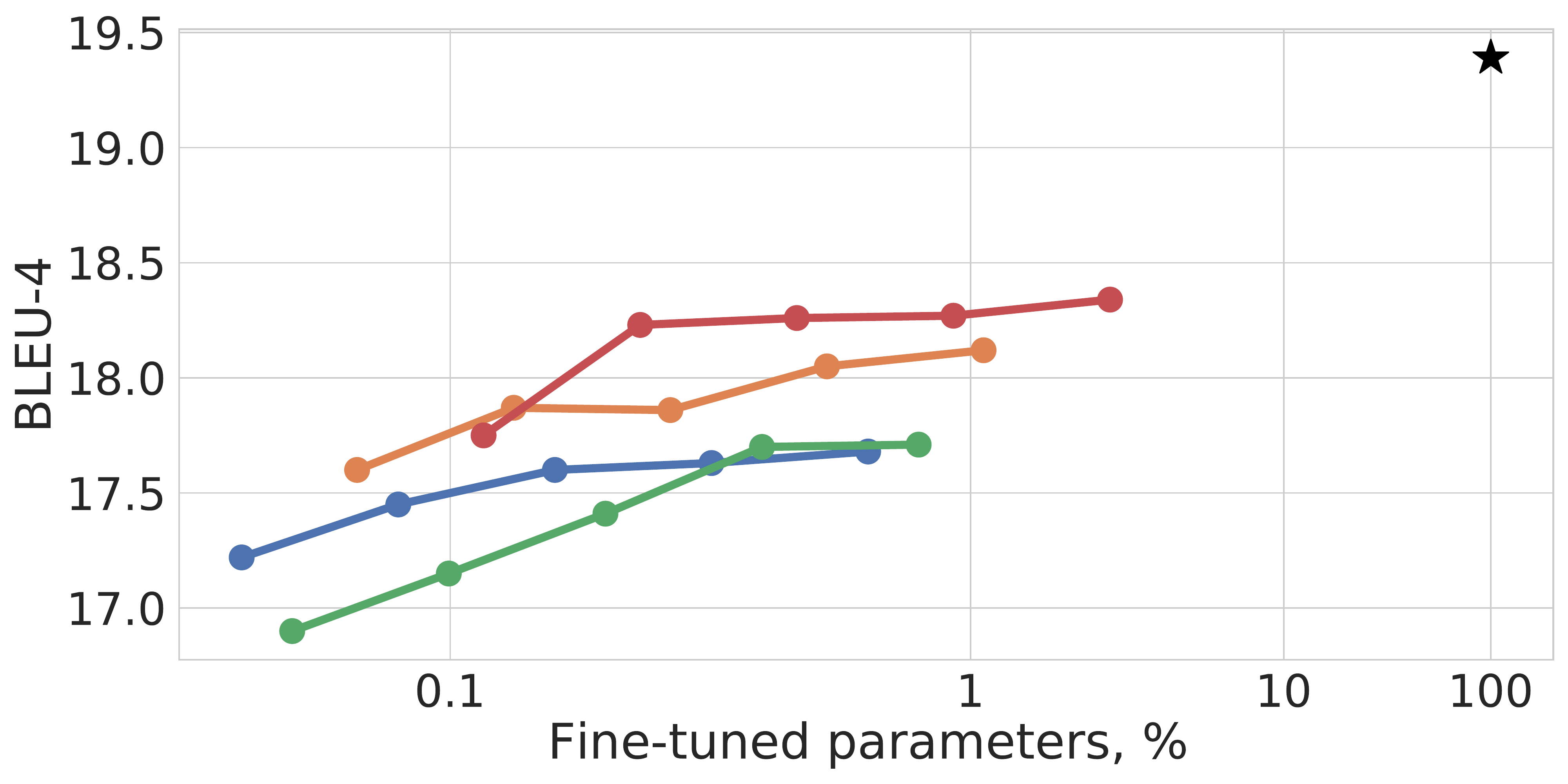}} &
        \subfloat[PLBART: code translation (C\#$\rightarrow$Java) \label{fig:add:plbart_transcj}]{\includegraphics[width=0.3\textwidth]{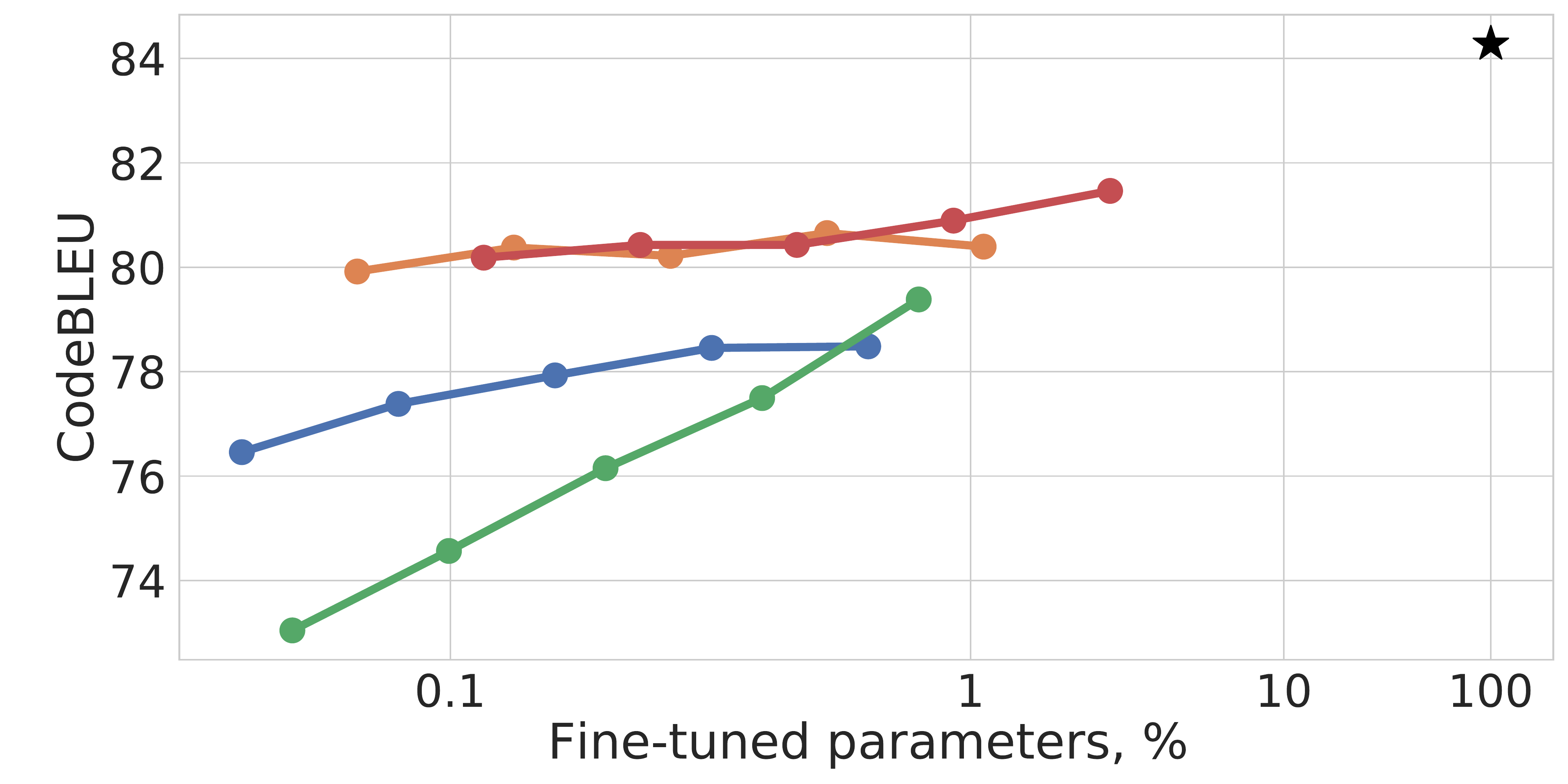}} \\
    \end{tabular}
    \caption{Quality of the PE fine-tuning methods in additional tasks on the test-set vs. the number of trainable parameters}
    \label{fig:add_results}
\end{figure*}

\begin{figure}
    \centering
    \includegraphics[width=0.55\textwidth]{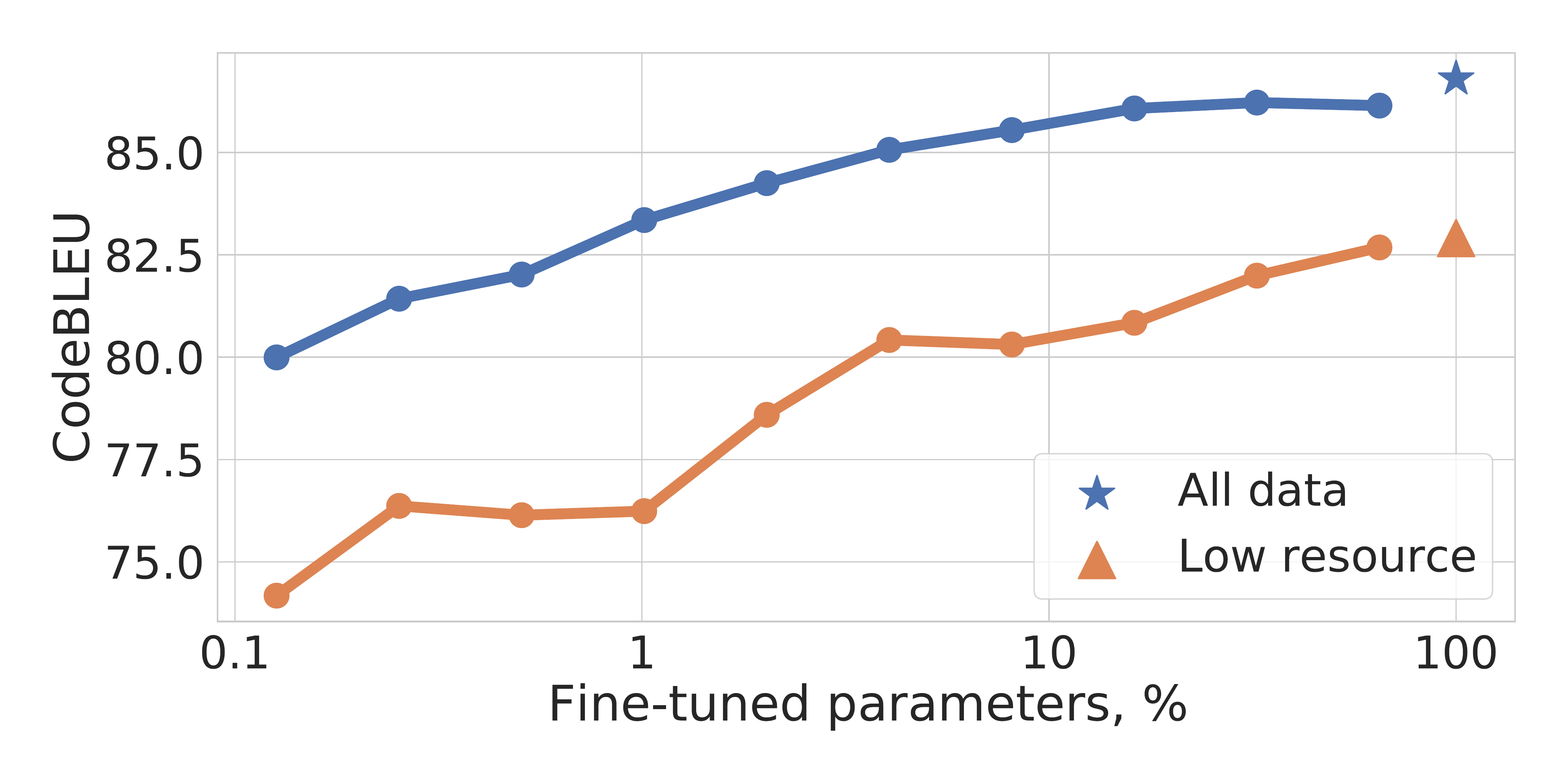}
    \caption{Quality of the Full fine-tuning (star/triangle) and FF-LoRA + Adapter (solid line) Training method when training on all data (10.3k training examples) and in low-resource setting (2.5k training examples) of the Java $\rightarrow$ C\# translate task vs. the number of trainable parameters}
    \label{fig:low-resource}
\end{figure}

Figure \ref{fig:add_results} shows results on additional tasks: code summarization on the Go language and code translation in other direction (C\#$\rightarrow$Java). The PE methods behave similarly as in the main text. Figure \ref{fig:low-resource} shows results of the FF-LoRA + Adapter Training method in low-resource setting (2.5k training examples vs. 10.3k training examples in full training set) of the Java $\rightarrow$ C\# translate task. The method performs similarly relative to the full fine-tuning as when using all available data. Studying the effect of the methods in low-resource scenario (and potentially in few-shot setting) is one of possible research directions.

Tables \ref{table:sumpy}-\ref{table:translate-cj} show quantitative results from Figures~\ref{fig:main_results} and \ref{fig:add_results} in the numerical form. Hyperparameter $r$ (methods' internal size) for each task/method set was chosen based on the quality on the dev set.
Table \ref{table:sumpy_examples} shows more CodeT5 predictions in the code summarization (Python) task.


\begin{table*}
    \centering
    \begin{tabular}{|c|l|r|c|}
        \hline
        Model & Method & \# Parameters & BLEU-4 ($\uparrow$) \\
        \hline
        \multirow{5}{3.7em}{CodeT5} & FT (baseline) & 223M (100\%) & 20.16 $\pm$ 0.22 \\
        & LoRA ($r$=2) & 0.2M (0.1\%) & \textbf{20.36} $\pm$ 0.05 \\
        & FF-LoRA ($r$=4) & 0.7M (0.3\%) & \textbf{20.36} $\pm$ 0.03 \\
        & AT ($r$=16) & 2.2M (1\%) & \textbf{20.36} $\pm$ 0.03 \\
        & FF-LoRA + AT ($r$ = 16) & 5.2M (2.3\%) & \textbf{20.36} $\pm$ 0.03 \\
        \hline
        \multirow{5}{3.7em}{PLBART} & FT (baseline) & 140M (100\%) & \textbf{19.63} $\pm$ 0.07 \\
        & LoRA ($r$=16) & 0.9M (0.6\%) & 19.23 $\pm$ 0.09 \\
        & FF-LoRA ($r$=16) & 1.5M (1\%) & 19.30 $\pm$ 0.12 \\
        & AT ($r$=16) & 1.1M (0.8\%) & 19.26 $\pm$ 0.05 \\
        & FF-LoRA + AT ($r$ = 8) & 1.3M (0.9\%) & 19.39 $\pm$ 0.03 \\
        \hline
    \end{tabular}
    \caption{PE methods quality on the test set in the \textbf{code summarization (Python)} task. Hyperparameter $r$ (methods' internal size) was chosen based on the quality on the dev set. Standard deviation is estimated in 3 runs.}
    \label{table:sumpy}
\end{table*}

\begin{table*}
    \centering
    \begin{tabular}{|c|l|r|c|c|c|}
        \hline
        Model & Method & \# Parameters & F1 ($\uparrow$) & Precision ($\uparrow$) & Recall ($\uparrow$) \\
        \hline
        \multirow{5}{3.7em}{CodeT5} & FT (baseline) & 224M (100\%) & \textbf{94.56} $\pm$ 0.28 & \textbf{95.74} $\pm$ 0.51 & 93.41 $\pm$ 0.15 \\
        & LoRA ($r$=4) & 1.6M (0.7\%) & 93.96 $\pm$ 0.46 & 93.26 $\pm$ 0.64 & 94.67 $\pm$ 0.40 \\
        & FF-LoRA ($r$=16) & 4.1M (1.8\%) & 93.66 $\pm$ 0.22 & 92.16 $\pm$ 0.36 & \textbf{95.21} $\pm$ 0.81 \\
        & AT ($r$=16) & 3.4M (1.5\%) & 93.76 $\pm$ 0.54 & 92.56 $\pm$ 0.76 & 94.99 $\pm$ 0.66 \\
        & FF-LoRA + AT ($r$=16) &  6.3M (2.8\%) & 93.91 $\pm$ 0.28 & 93.83 $\pm$ 0.61 & 93.99 $\pm$ 0.48 \\
        \hline
        \multirow{5}{3.7em}{PLBART} & FT (baseline) & 141M (100\%) & \textbf{94.03} $\pm$ 0.33 & \textbf{94.00} $\pm$ 0.62 & 94.06 $\pm$ 0.54 \\
        & LoRA ($r$=1) & 1.2M (0.9\%) & 92.89 $\pm$ 0.24 & 91.40 $\pm$ 0.24 & 94.43 $\pm$ 0.33 \\
        & FF-LoRA ($r$=16) & 2.7M (1.9\%) & 93.75 $\pm$ 0.24 & 92.91 $\pm$ 0.90 & 94.63 $\pm$ 0.56 \\
        & AT ($r$=8) & 1.7M (1.2\%) & 92.49 $\pm$ 0.29 & 90.77 $\pm$ 0.31 & 94.28 $\pm$ 0.43 \\
        & FF-LoRA + AT ($r$=2) &  1.5M (1.1\%) & 93.62 $\pm$ 0.03 & 92.32 $\pm$ 0.24 & \textbf{94.96} $\pm$ 0.30 \\
        \hline
    \end{tabular}
    \caption{PE methods quality on the test set in the \textbf{code clone detection} task. Hyperparameter $r$ (methods' internal size) was chosen based on the quality on the dev set. Standard deviation is estimated in 3 runs.}
    \label{table:clone}
\end{table*}

\begin{table*}
    \centering
    \begin{tabular}{|c|l|r|c|c|c|}
        \hline
        Model & Method & \# Parameters & BLEU-4 ($\uparrow$) & EM ($\uparrow$) & CodeBLEU($\uparrow$) \\
        \hline
        \multirow{5}{3.6em}{CodeT5} & FT (baseline) & 223M (100\%) & \textbf{39.79} $\pm$ 0.99 & \textbf{22.33} $\pm$ 0.34 & \textbf{42.95} $\pm$ 0.91 \\
        & LoRA ($r$=512) & 57M (25.4\%) & 34.80 $\pm$ 0.12 & 21.52 $\pm$ 0.28 & 39.01 $\pm$ 0.11 \\
        & FF-LoRA ($r$=256) & 47M (21.2\%) & 36.17 $\pm$ 0.10 & 21.65 $\pm$ 0.15 & 39.72 $\pm$ 0.13 \\
        & AT ($r$=512) & 71M (31.8\%) & 38.25 $\pm$ 0.42 & 21.92 $\pm$ 0.37 & 41.78 $\pm$ 0.16 \\
        & FF-LoRA + AT ($r$=512) &  165M (74\%) & 39.45 $\pm$ 0.58 & 22.08 $\pm$ 0.19 & 42.71 $\pm$ 0.26 \\
        \hline
        \multirow{5}{3.6em}{PLBART} & FT (baseline) & 140M (100\%) & \textbf{39.40} $\pm$ 1.05 & \textbf{21.08} $\pm$ 0.21 & \textbf{42.64} $\pm$ 0.63 \\
        & LoRA ($r$=8) & 0.4M (0.3\%) & 26.45 $\pm$ 0.30 & 19.20 $\pm$ 0.04 & 34.91 $\pm$ 0.15 \\
        & FF-LoRA ($r$=128) & 12M (8.5\%) & 29.28 $\pm$ 0.37 & 20.13 $\pm$ 0.09 & 36.87 $\pm$ 0.26 \\
        & AT ($r$=512) & 35M (25.4\%) & 32.57 $\pm$ 0.40 & 20.47 $\pm$ 0.15 & 39.86 $\pm$ 0.31 \\
        & FF-LoRA + AT ($r$=512) &  83M (59.3\%) & 33.80 $\pm$ 0.47 & 20.68 $\pm$ 0.06 & 39.79 $\pm$ 0.50 \\
        \hline
    \end{tabular}
    \caption{PE methods quality on the test set in the \textbf{code generation} task. Hyperparameter $r$ (methods' internal size) was chosen based on the quality on the dev set. Standard deviation is estimated in 3 runs.}
    \label{table:gen}
\end{table*}

\begin{table*}
    \centering
    \begin{tabular}{|c|l|r|c|c|c|}
        \hline
        Model & Method & \# Parameters & BLEU-4 ($\uparrow$) & EM ($\uparrow$) & CodeBLEU($\uparrow$) \\
        \hline
        \multirow{5}{3.7em}{CodeT5} & FT (baseline) & 223M (100\%) & \textbf{83.42} $\pm$ 0.42 & \textbf{64.63} $\pm$ 1.10 & \textbf{86.80} $\pm$ 0.32 \\
        & LoRA ($r$=64) & 7.1M (3.2\%) & 76.30 $\pm$ 0.22 & 54.43 $\pm$ 0.29 & 81.22 $\pm$ 0.19 \\
        & FF-LoRA ($r$=64) & 11.8M (5.3\%) & 78.23 $\pm$ 0.09 & 55.73 $\pm$ 0.26 & 82.38 $\pm$ 0.13 \\
        & AT ($r$=512) & 71M (31.8\%) & 82.92 $\pm$ 0.07 & 63.73 $\pm$ 0.62 & 86.56 $\pm$ 0.04 \\
        & FF-LoRA + AT ($r$=512) &  165M (74\%) & 82.39 $\pm$ 0.45 & 63.03 $\pm$ 0.49 & 86.09 $\pm$ 0.35 \\
        \hline
        \multirow{5}{3.7em}{PLBART} & FT (baseline) & 140M (100\%) & \textbf{81.85} $\pm$ 0.33 & \textbf{59.50} $\pm$ 0.99 & \textbf{85.50} $\pm$ 0.20 \\
        & LoRA ($r$=512) & 28M (20.3\%) & 71.86 $\pm$ 0.02 & 47.67 $\pm$ 0.09 & 77.75 $\pm$ 0.07 \\
        & FF-LoRA ($r$=32) & 3M (2.1\%) & 75.16 $\pm$ 0.12 & 51.23 $\pm$ 0.12 & 80.18 $\pm$ 0.08 \\
        & AT ($r$=512) & 35M (25.4\%) & 80.22 $\pm$ 0.40 & 56.10 $\pm$ 0.78 & 84.19 $\pm$ 0.37 \\
        & FF-LoRA + AT ($r$=512) &  83M (59.3\%) & 80.42 $\pm$ 0.12 & 56.90 $\pm$ 0.49 & 84.36 $\pm$ 0.16 \\
        \hline
    \end{tabular}
    \caption{PE methods quality on the test set in the \textbf{code translation (Java $\rightarrow$ C\#)} task. Hyperparameter $r$ (methods' internal size) was chosen based on the quality on the dev set. Standard deviation is estimated in 3 runs.}
    \label{table:translate}
\end{table*}

\begin{table*}
    \centering
    \begin{tabular}{|c|l|r|c|}
        \hline
        Model & Method & \# Parameters & BLEU-4 ($\uparrow$) \\
        \hline
        \multirow{5}{3.7em}{CodeT5} & FT (baseline) & 223M (100\%) & \textbf{19.68} $\pm$ 0.17 \\
        & LoRA ($r$=8) & 0.9M (0.4\%) & 19.47 $\pm$ 0.00 \\
        & FF-LoRA ($r$=8) & 1.5M (0.7\%) & 19.41 $\pm$ 0.29 \\
        & AT ($r$=8) & 1.1M (0.5\%) & 19.12 $\pm$ 0.11 \\
        & FF-LoRA + AT ($r$=2) & 0.6M (0.3\%) & 19.17 $\pm$ 0.25 \\
        \hline
        \multirow{5}{3.7em}{PLBART} & FT (baseline) & 140M (100\%) & \textbf{19.11} $\pm$ 0.19 \\
        & LoRA ($r$=16) & 0.9M (0.6\%) & 18.32 $\pm$ 0.04 \\
        & FF-LoRA ($r$=16) & 1.5M (1.1\%) & 18.38 $\pm$ 0.04 \\
        & AT ($r$=16) & 1.1M (0.8\%) & 18.03 $\pm$ 0.04 \\
        & FF-LoRA + AT ($r$=4) & 0.6M (0.5\%) & 18.40 $\pm$ 0.12 \\
        \hline
    \end{tabular}
    \caption{PE methods quality on the test set in the \textbf{code summarization (Go)} task. Hyperparameter $r$ (methods' internal size) was chosen based on the quality on the dev set. Standard deviation is estimated in 3 runs.}
    \label{table:sumgo}
\end{table*}

\begin{table*}
    \centering
    \begin{tabular}{|c|l|r|c|}
        \hline
        Model & Method & \# Parameters & BLEU-4 ($\uparrow$) \\
        \hline
        \multirow{5}{3.7em}{CodeT5} & FT (baseline) & 223M (100\%) & \textbf{20.23} $\pm$ 0.09 \\
        & LoRA ($r$=16) & 1.8M (0.8\%) & 20.04 $\pm$ 0.02 \\
        & FF-LoRA ($r$=16) & 3M (1.3\%) & 20.04 $\pm$ 0.04 \\
        & AT ($r$=16) & 2.2M (1\%) & 20.01 $\pm$ 0.04 \\
        & FF-LoRA + AT ($r$=16) & 5.2M (2.3\%) & 20.15 $\pm$ 0.06 \\
        \hline
        \multirow{5}{3.7em}{PLBART} & FT (baseline) & 140M (100\%) & \textbf{19.51} $\pm$ 0.11 \\
        & LoRA ($r$=16) & 0.9M (0.6\%) & 17.32 $\pm$ 0.48 \\
        & FF-LoRA ($r$=8) & 0.7M (0.5\%) & 18.10 $\pm$ 0.06 \\
        & AT ($r$=16) & 1.1M (0.8\%) & 17.78 $\pm$ 0.05 \\
        & FF-LoRA + AT ($r$=16) & 2.6M (1.8\%) & 18.40 $\pm$ 0.08 \\
        \hline
    \end{tabular}
    \caption{PE methods quality on the test set in the \textbf{code summarization (Java)} task. Hyperparameter $r$ (methods' internal size) was chosen based on the quality on the dev set. Standard deviation is estimated in 3 runs.}
    \label{table:sumjava}
\end{table*}

\begin{table*}
    \centering
    \begin{tabular}{|c|l|r|c|c|c|}
        \hline
        Model & Method & \# Parameters & BLEU-4 ($\uparrow$) & EM ($\uparrow$) & CodeBLEU($\uparrow$) \\
        \hline
        \multirow{5}{3.7em}{CodeT5} & FT (baseline) & 223M (100\%) & \textbf{79.14} $\pm$ 0.30 & \textbf{66.23} $\pm$ 0.29 & \textbf{84.33} $\pm$ 0.25 \\
        & LoRA ($r$=8) & 0.9M (0.4\%) & 74.26 $\pm$ 0.12 & 59.27 $\pm$ 0.24 & 80.38 $\pm$ 0.12 \\
        & FF-LoRA ($r$=8) & 1.5M (0.7\%) & 75.34 $\pm$ 0.17 & 61.00 $\pm$ 0.22 & 81.08 $\pm$ 0.13 \\
        & AT ($r$=16) & 2.2M (1\%) & 75.30 $\pm$ 0.31 & 60.43 $\pm$ 0.37 & 81.08 $\pm$ 0.13 \\
        & FF-LoRA + AT ($r$=16) &  5.2M (2.3\%) & 76.89 $\pm$ 0.10 & 62.40 $\pm$ 0.22 & 82.53 $\pm$ 0.15 \\
        \hline
        \multirow{5}{3.7em}{PLBART} & FT (baseline) & 140M (100\%) & \textbf{78.60} $\pm$ 0.21 & \textbf{64.70} $\pm$ 0.08 & \textbf{84.12} $\pm$ 0.13 \\
        & LoRA ($r$=16) & 0.9M (0.6\%) & 71.20 $\pm$ 0.03 & 55.10 $\pm$ 0.16 & 78.48 $\pm$ 0.02 \\
        & FF-LoRA ($r$=16) & 1.5M (1.1\%) & 74.27 $\pm$ 0.27 & 59.10 $\pm$ 0.22 & 80.59 $\pm$ 0.14 \\
        & AT ($r$=16) & 1.1M (0.8\%) & 72.02 $\pm$ 0.20 & 55.30 $\pm$ 0.29 & 79.18 $\pm$ 0.18 \\
        & FF-LoRA + AT ($r$=16) &  2.6M (1.9\%) & 75.11 $\pm$ 0.25 & 59.87 $\pm$ 0.19 & 81.22 $\pm$ 0.17 \\
        \hline
    \end{tabular}
    \caption{PE methods quality on the test set in the \textbf{code translation (C\# $\rightarrow$ Java)} task. Hyperparameter $r$ (methods' internal size) was chosen based on the quality on the dev set. Standard deviation is estimated in 3 runs.}
    \label{table:translate-cj}
\end{table*}

\begin{table*}
    \small
    \centering
    \begin{tabular}{|l|l|p{13cm}|}
        \hline
        1 & GT & str - > list Convert XML to URL List . From Biligrab . \\*
        {} & FT & Convert Sina XML to a list of URLs . \\*
        {} & LoRA & Convert Sina XML to a list of URLs . \\*
        {} & FF-LoRA & Convert Sina XML to a list of URLs . \\*
        {} & AT & Convert an XML string to a list of URLs . \\
        \hline
        2 & GT & Print a log message to standard error . \\*
        {} & FT & Print a log message to stderr . \\*
        {} & LoRA & Print a log message to stderr . \\*
        {} & FF-LoRA & Print a log message to stderr . \\*
        {} & AT & Print a log message to stderr . \\
        \hline
        3 & GT & What a Terrible Failure! \\*
        {} & FT & Print a message to stdout and exit . \\*
        {} & LoRA & Print a message to the console . \\*
        {} & FF-LoRA & Print a message to the console . \\*
        {} & AT & Print a message to the console . \\
        \hline
        4 & GT & Detect operating system . \\*
        {} & FT & Detect OS . \\*
        {} & LoRA & Detect the operating system . \\*
        {} & FF-LoRA & Detects the operating system . \\*
        {} & AT & Detect the operating system . \\
        \hline
        5 & GT & Converts a string to a valid filename . \\*
        {} & FT & Return text legitimately according to OS . \\*
        {} & LoRA & Convert a string to a legitimate string . \\*
        {} & FF-LoRA & Turn a string into a legitimate string . \\*
        {} & AT & Turn a string into a legitimate string . \\
        \hline
        6 & GT & Override the original one Ugly ugly dirty hack \\*
        {} & FT & Download the stream . \\*
        {} & LoRA & Download a stream . \\*
        {} & FF-LoRA & Download a stream . \\*
        {} & AT & Download a stream . \\
        \hline
        7 & GT & Scans through a string for substrings matched some patterns . \\*
        {} & FT & matchall - Searches text for all patterns in patterns and returns a list of matches \\*
        {} & LoRA & Return a list of all matching patterns in text . \\*
        {} & FF-LoRA & Match all patterns in text . \\*
        {} & AT & Match all patterns in text . \\
        \hline
        8 & GT & JSON int int int - > str Get a proper title with courseid + topicID + partID . \\*
        {} & FT & Get the title of a specific topic and part by its index . \\*
        {} & LoRA & Get the title of a topic and part . \\*
        {} & FF-LoRA & Get the title of a topic part . \\*
        {} & AT & Get the title for a given topic and part . \\
        \hline
        9 & GT & int - > None Download a WHOLE course . Reuse the API call to save time . \\*
        {} & FT & wanmen\_download\_by\_course\_topic and wanmen\_download\_by\_course\_part recursively \\*
        {} & LoRA & Download wanmen by course . \\*
        {} & FF-LoRA & Download wanmen files by course . \\*
        {} & AT & Download wanmen by course . \\
        \hline
        10 & GT & Returns and flush the event buffer . In case dag\_ids is specified it will only return and flush events for the given dag\_ids . Otherwise it returns and flushes all \\*
        {} & FT & Clear the event buffer for the given DAGs . If no DAGs are given clear the event buffer . \\*
        {} & LoRA & Get the event buffer for the given dag ids . \\*
        {} & FF-LoRA & Get the event buffer for the given dag\_ids . \\*
        {} & AT & Get the event buffer for the given dag\_ids . \\
        \hline
    \end{tabular}
\caption{CodeT5 predictions in code summarization (Python) task when using different PE methods. GT stands for Ground Truth}
\label{table:sumpy_examples}
\end{table*}

\end{document}